\documentclass{ieeeaccess}
\usepackage{amsmath,amssymb,amsfonts}
\usepackage{algorithmic}
\usepackage{graphicx}
\usepackage{textcomp}
\usepackage[numbers]{natbib}
\usepackage{makecell}
\renewcommand{\textsc}[1]{\texttt{#1}}

\usepackage{bm}
\makeatletter
\AtBeginDocument{\DeclareMathVersion{bold}
\SetSymbolFont{operators}{bold}{T1}{times}{b}{n}
\SetSymbolFont{NewLetters}{bold}{T1}{times}{b}{it}
\SetMathAlphabet{\mathrm}{bold}{T1}{times}{b}{n}
\SetMathAlphabet{\mathit}{bold}{T1}{times}{b}{it}
\SetMathAlphabet{\mathbf}{bold}{T1}{times}{b}{n}
\SetMathAlphabet{\mathtt}{bold}{OT1}{pcr}{b}{n}
\SetSymbolFont{symbols}{bold}{OMS}{cmsy}{b}{n}

\newcommand{\norm}[1]{\left\lVert#1\right\rVert}
\newcommand{\matr}[1]{\mathbf{#1}}

\usepackage{tabulary}
\usepackage{tabularx}

\usepackage{multirow}
\renewcommand\boldmath{\@nomath\boldmath\mathversion{bold}}}
\makeatother

\def\BibTeX{{\rm B\kern-.05em{\sc i\kern-.025em b}\kern-.08em
    T\kern-.1667em\lower.7ex\hbox{E}\kern-.125emX}}

\begin{document}
\history{Date of publication xxxx 00, 0000, date of current version xxxx 00, 0000.}
\doi{10.1109/ACCESS.2023.1120000}

\title{Injecting linguistic knowledge into BERT for Dialogue State Tracking}
\author{\uppercase{Xiaohan Feng}\authorrefmark{1}, 
\uppercase{Xixin Wu}\authorrefmark{2}\IEEEmembership{Member, IEEE}, and \uppercase{Helen Meng}\authorrefmark{3}\IEEEmembership{Fellow, IEEE}
}

\address[1,2,3]{Department of System Engineering and Engineering Management, the Chinese University of Hong Kong  (e-mail: \{xhfeng, wuxx, hmmeng\}@se.cuhk.edu.hk)}
\tfootnote{This work is partially supported by the General Research Fund from the Research Grants Council of Hong Kong SAR Government (Project No. 14207619).}

\markboth
{Author \headeretal: Preparation of Papers for IEEE Access}
{Author \headeretal: Preparation of Papers for IEEE Access}

\corresp{Corresponding author: Helen Meng (e-mail: hmmeng@se.cuhk.edu.hk).}

\begin{abstract}

Dialogue State Tracking (DST) models often employ intricate neural network architectures, necessitating substantial training data, and their inference process lacks transparency. This paper proposes a method that extracts linguistic knowledge via an unsupervised framework and subsequently utilizes this knowledge to augment BERT's performance and interpretability in DST tasks. The knowledge extraction procedure is computationally economical and does not require annotations or additional training data. The injection of the extracted knowledge can be achieved by the addition of simple neural modules. We employ the Convex Polytopic Model (CPM) as a feature extraction tool for DST tasks and illustrate that the acquired features correlate with syntactic and semantic patterns in the dialogues. This correlation facilitates a comprehensive understanding of the linguistic features influencing the DST model’s decision-making process. We benchmark this framework on various DST tasks and observe a notable improvement in accuracy.
\end{abstract}

\begin{keywords}
Dialogue State Tracking; Convex Polytopic Model; Knowledge Extraction; Interpretable AI
\end{keywords}

\titlepgskip=-21pt

\maketitle

\section{Introduction}

\PARstart{D}{ialogue} State Tracking (DST) is a crucial component in task-oriented dialogue systems. DST aims to extract and update current dialogue state, which is typically represented as a set of slot-value pairs that capture the user's intent and the key information provided so far in the conversation. For example, in a flight booking system, given the user query \textbf{"What flights are there on Sunday from Seattle to Chicago?"}, the corresponding slot-value pairs could include \textsc{departure-city}: \textit{Seattle}, \textsc{arrival-city}: \textit{Chicago}, and \textsc{departure-date}: \textit{Sunday}. The dialogue state is used to generate appropriate system responses to fulfill the user's request. As such, accurate DST is essential for natural language understanding and successful task completion in conversational AI applications.

While neural network models, particularly those employing transformer-based architectures like BERT \cite{DBLP:journals/corr/abs-1810-04805}, have been reported to have achieved substantial performance enhancements in DST \citep{balaraman_recent_2021}, surpassing traditional heuristic-based approaches, we note that they also present challenges. These models, due to their intricate architectures, offer limited visibility and control during training and inference process. The cascading non-linear transformations from dialogue data to dialogue state are intricate and obscure. The latent representation space used in their intermediate layers may not align with the syntactical or semantic features in the input \cite{dalvi2022discovering}, leading to insufficient transparency. On the performance improvement side, the predominant strategies have been increasingly larger, task-specific datasets or to apply complex training paradigms to limited data samples \cite{jacqmin2022do}.

\begin{figure}[htp]
    \centering
    \includegraphics[width=\columnwidth]{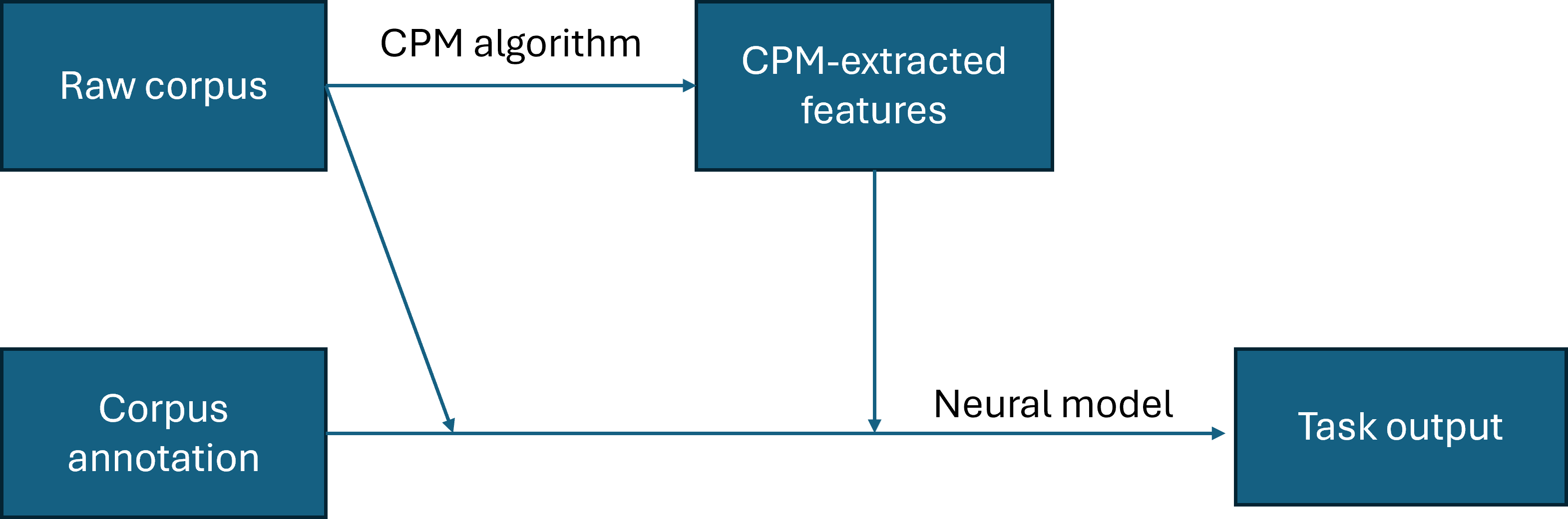}
    \caption{Flowchart of our proposed pipeline.}
    \label{fig:flowchart}
\end{figure}

Given these circumstances, this paper explores a novel approach. We aim to guide the training process with externally acquired knowledge, obtaining which does not require explicit annotations, while concurrently minimizing the need for additional computational resources beyond training the original model. This methodology seeks to infuse an interpretable feature set into neural models' opaque decision process, allowing analysis of the decision-making process by calculating the attributions to the features. A summary of the process is described in the flowchart in Figure \ref{fig:flowchart}.







This paper introduces a framework that leverages the Convex Polytopic Model (CPM) for semantic feature extraction in DST, offering computational efficiency and robustness. CPM has proven its efficacy in topic discovery tasks \cite{wu_topic_2018}, efficiently classifying documents into interpretable topics, even within short-text corpora, aligning with the concise nature of dialogue processing tasks. 

In contrast to neural-model-based representation models, feature extraction using CPM can efficiently run on CPUs during both training and inference stages, reducing computational resource burdens. The patterns discovered by CPM without supervision, are correlated to various intents and slots prevalent in task-oriented dialogues \cite{zhou_automatic_2021}, \cite{zhou_convex_2022}. These discovered patterns, easily interpretable by humans, enrich the approach's overall comprehensibility.

This study aims to illustrate the efficacy of CPM in DST models to improve both accuracy and interpretability compared to traditional neural-model based DST approaches. We first describe how to process dialogue datasets using our CPM-based framework, examining semantic structures and the relations of CPM features to dialogue states from a semantics viewpoint, to demonstrate that CPM can extract semantically relevant features without supervision. Subsequently, we show these semantic features are incorporated into the BERT architecture for DST, and then validate the improvements in DST by fine-tuning two pre-trained models—one incorporating CPM features and the other not—on DST datasets, showing that the inclusion of CPM features enables the BERT-based model to achieve superior scores in DST tasks without necessitating additional fine-tuning strategies or data samples. Finally, we visualize and interpret the CPM's impact on the DST model by calculating the attribution of key tokens to accurate DST prediction. 
We use this work as the first step of using CPM on a series of dialogue processing tasks involving neural models which lack interpretability, and we examine the effect on a BERT-based model as a demonstration of the framework, as well as a clear visualization of interpretable features from CPM influencing the decision from neural model.

Our contributions can be summarized as follows:
\begin{enumerate}
    \item We pioneer the application of CPM to DST, employing it as semantic context in the context-guided BERT model.
    \item We design a CPM-guided BERT structure that surpasses baseline performance on DST task.
    \item We assess the effect of CPM-extracted features on the BERT model, underscoring CPM’s role in enhancing the interpretability of neural-model-based DST systems.
\end{enumerate}

\section{Background}

\textbf{Dialogue State Tracking}. DST models have evolved from employing parsing rules, to pinpointing dialogue state slot-value pairs within dialogues to leveraging neural models, to encoding dialogues as latent representations \cite{wen_network-based_2017}, \cite{mrksic_neural_2017-2}. Recent advancements in DST models predominantly utilize transformer-based encoders \cite{lai_simple_2020}, \cite{hosseini-asl_simple_2020}. For decoders, the strategy includes discriminating decoders \cite{chao_bert-dst_2019}, \cite{lee_sumbt_2019}, \cite{heck_trippy_2020} and generation-based models \cite{feng_sequence--sequence_2021}. 

Given the data quantity requirement from transformer encoders, numerous works have focused on optimizing decoding/generation strategies, such as pointer network \cite{mrksic_neural_2017-2}, and grounding responses using external context data like database query results \cite{peng2021soloist}. Others have aimed at augmenting and denoising data, synthesizing annotations \cite{heck_robust_2022}, and translating dialogue queries to SQL \cite{hu_-context_2022}. 

Our work uniquely centers on the encoder module of the DST pipeline, often overshadowed in research due to the intricacy and opacity of PLMs. Even as this paper utilizes a relatively straightforward baseline model to underscore the efficacy of our approach, it can seamlessly integrate into diverse works, fostering more robust and performant systems.

\textbf{Linguistic knowledge and PLM.}
Transformer-based Pre-trained Language Models (PLM) are known for proficiently modeling inter-token dependencies, primarily covered by linguistic information domains, through the self-attention mechanism \cite{vaswani2017attention}.
Some research works have demonstrated that transformer layers score well in synthetic dependency tasks \cite{clark_what_2019-1}, and conjectured that and that this sensitivity to dependency relationships is pivotal for downstream NLP tasks \cite{clark_what_2019-1}, \cite{rogers_primer_2020}. However, this dependency relationship is not explicitly utilized to guide the pre-training and fine-tuning of PLMs, nor do PLMs produce any explicit intermediate representation of dependencies or other linguistic features. Hence, studies have explored the injection of contextual information into the self-attention mechanism, improving tasks like machine translation \cite{yang_context-aware_2019} and sentiment analysis \cite{wu_context-guided_2020}.

These studies, for simplicity, often avoided using external contextual knowledge, instead, relying on transformations of latent representations in PLMs as sources of contextual knowledge. Although these latent representations are context-aware, they remain susceptible to interpretability issues. Some solutions, such as using a specialized neural component to extract contextualized word representations \cite{peters-etal-2018-deep}, \cite{liu-etal-2019-linguistic}, require additional fine-tuning.

In contrast, our approach employs a computationally efficient algorithm to capture semantic patterns in the entire training corpus and generates features that guide attention to semantically important tokens in the input sequence. This process and the resulting features can be easily analyzed and understood.

\section{Model}

\subsection{Task: Dialogue State Tracking}

Dialogue State Tracking (DST) involves receiving the dialogue history, the current dialogue turn, and dialogue state history as inputs and yields the updated dialogue state for the current turn. More precisely, the dialogue state often adopts the format of \textit{domain-slot: value}. Here, the slot serves as a domain-specific tag, attributing the specified value. Depending on the nature of the slot, the values may fall within a pre-defined category set, for instance, a binary choice of "yes" or "no" for the slot \textit{hotel-internet}. It is noteworthy that in the same dialogue turn, multiple slots across various domains may undergo updates.

\subsection{Base model: TripPy}

We build our model on the foundational principles laid out by Heck \cite{heck_trippy_2020}, using their well-regarded dialogue tracking model, TripPy, as our point of departure and baseline for modifications. TripPy sources slot values from three pivotal points: user utterances, system-informed values (current or preceding turns), and values derived from other slots. It divides the prediction of dialogue states into two subtasks: discerning the slots requiring attention in the ongoing turn and forecasting values for all slots slated for updates.

Given the preceding dialogue state, an update decision tag—\textit{{none, dontcare, span, inform, refer}}—is anticipated for all non-boolean slots in the schema. For boolean slots, the set \textit{{none, dontcare, yes, no}} is used. The tag \textit{none} implies no updates to the correlated slot, while  \textit{dontcare, yes, no} denote actions updating the slot value per the tag. The \textit{span} indicates an update using current user turn values by pinpointing the start and end positions of the value to be replicated; \textit{inform} means that the update value should be from previously system-informed memories, and \textit{refer} implies value-sharing between slots.

This DST pipeline is implemented utilizing an encoder-only BERT architecture, with distinct classifiers leveraging the latent representation from the encoder to predict slot actions and values. The input is comprised of current user and system turns and the prior dialogue state, which are concatenated and encoded via a single BERT model. The slot action is anticipated by a linear classifier on the representation of the first token "[CLS]" , with a distinct classifier being trained for each slot defined in the ontology. The span of slot value in the input is deduced using a pair of classifiers on all other tokens for each slot, applying a Softmax to the output score to formulate a probability distribution for the start and end positions of the desired value in the input. This mechanism remains constant for informed memory and dialogue state memory, where the classifier selects one value from candidate lists.

As a high-performing model, TripPy has gathered considerable acclaim in recent DST literature. However, its latent representation from the encoder remains intricate and challenging to decipher. We address this limitation by integrating a set of interpretable features into BERT's attention layer in our model, aligning them with the semantic properties of input dialogue to influence the DST model's training and inference processes. The following sections will elaborate on the interpretability of these features.

\subsection{Knowledge extraction: CPM}
\label{sec:CPM}
This section gives a brief introduction of Convex Polytopic Model (CPM), a notable algorithm capable of extracting pivotal concepts within a corpus. These concepts have been discovered to correlate with slots and values in task-oriented dialogues. For an in-depth exploration of CPM, readers are directed to our preceding publications.

CPM initializes with utterances depicted through a bag-of-words model, with each utterance symbolized by a word frequency vector. This vector’s dimension, $F$  is determined by the vocabulary size and is sum-normalized across columns. Subsequently, Principal Component Analysis (PCA) is employed to condense these vectors into a lower dimension, $R$. Thereupon, a convex polytope is formulated by identifying the minimal volume of the polytope that envelops all $R$-dim vectors with $R+1$ vertices, labelled as $v_1, \ldots, v_{R+1}$, utilizing the algorithm proposed by \cite{li_minimum_2008}.

This transformation yields two valuable geometric features:

\begin{enumerate}
    \item \textbf{Composition coefficients of vertices} with respect to utterances: any representations of the utterances $p_i$ in $R$-dim space can be decomposed in the format of the following:
    \begin{equation}
        p_i = \sum^K_{j=1} a_{ij} v_j
    \end{equation}
    where $a_{ij}$ are all non-negative, with a higher value signifying a stronger association between the vertex and the utterance. The composition coefficients may be directly used as a feature related to semantic frame in the utterance, while the vertex coefficients could be utilized with the following transformations: 

    \item \textbf{Vertex coordinates of the polytope}: the coordinates of vertices in the reduced $R$-dim subspace can be projected back to the $N$-dim space, and in each $N$-dim vector of vertex, words with the highest weight can be considered the most correlated words for the vertex. 
    Stacking all vertex coordinate vectors horizontally renders a vertex matrix $\matr{V} \in \mathbb{R}^{N \times (R+1)}$, wherein each row corresponds to a vocabulary word. Subsequently, every entry in a row serves as the correlation between the vertex and the corresponding column and word, 
enabling each word to be represented by its row vector $\matr{V}^{T}_{i}{\in K}$, This leads to the derivation of a word-word correlation matrix $\matr{M}$ for each utterance through cosine similarity on word vectors:
\begin{equation}
    \matr{M}_{[i,j]} = \frac{\matr{V}^{T}_{i} \cdot \matr{V}^{T}_{j}}{\norm{\matr{V}^{T}_{i}} \norm{\matr{V}^{T}_{j}}}
    \label{eq:coss}
\end{equation}
where $\matr{M}_{[i,j]}$ denotes the similarity between $i$-th and $j$-th words in the utterance. For example, in the utterance \textit{"All flights from Denver to Pittsburgh leaving after 6pm and before 7pm"}, the tokens within "after 6pm and before 7pm" exhibit substantial correlations among themselves, contrasting their weaker relations to the external tokens like "Pittsburgh" and "leaving".
\end{enumerate}

In essence, CPM features encapsulate crucial semantic patterns, including phrases encompassing slots and values, which surpass the limitation of linear word order, and often embodying hierarchical structures. The composition coefficients can be directly correlated to the semantic frame in the utterance, while the vertex coefficients can undergo further transformations.

The interpretability of CPM originates from its foundation in the bag-of-words subspace, enabling geometric properties in the reduced subspace to revert to the original bag-of-words space, interpretable through the relationship among the set of basis, corresponding to the corpus vocabulary. To illustrate this interpretability, we process all dialogue turn pairs solely related to the \textit{train} domain in the MultiWoZ 2.4 dataset, utilizing WordPiece tokenizer. Words with less than two occurrences were substituted with a unique "[UNK]" token, and a 3-dimensional MVS encompassing all representation points of dialogue turns in the reduced 3-D subspace is calculated.

\begin{figure}[htp]
    \centering
    \includegraphics[width=\columnwidth]{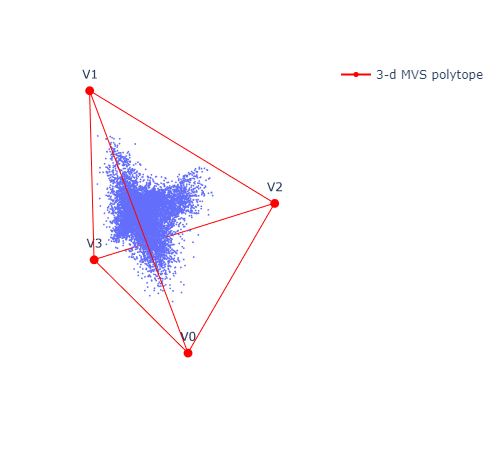}
    \caption{The 3-D MVS-type polytope. Vertices(Red and in bold) are labelled as V1-V4. The scattered dots denote projected utterance points.}
    \label{fig:3d-mvs}
\end{figure}

\begin{table*}[htp]
    \centering
\begin{tabular}{c c}
        Vertex & Nearest 3 utterances and Top-10 words \\
        \multirow{4}{*}{V1} & … i want one that \textit{leaves after 13:00}  \\
        &  … can I have the \textit{travel time}?…  \\
        & The \textit{9:59 train} leaves cambridge …  \\
        \cline{2-2}
        & [UNK],at,that,.,the,tr,for,book,leaves,would\\
        \hline\hline
        \multirow{4}{*}{V2} & \textbf{where will you be departing from?  i will be departing from} ely  \\
        &  \textbf{where will you be departing from?  i will be departing from} norwich  \\
        & \textbf{where will you be departing from?  i will be departing from} cambridge  \\
        \cline{2-2}
        & \textbf{from,be,will,?,you,depart,where},.,leavin,are\\
        \hline\hline
        \multirow{4}{*}{V3} & … \textbf{i need a train to} Cambridge  \\
        &  … \textbf{i need a train} on Wednesday \\
        & … \textbf{i need a train going to} Cambridge  \\
        \cline{2-2}
        & \textbf{train,a,i,.,need,to},cambri,for,lookin,on\\
        \hline\hline
        \multirow{4}{*}{V4} & … \textbf{i would like to go to} Cambridge  \\
        &  … \textbf{i would like to go to} ely please \\
        & … \textbf{i would like to leave} on Friday  \\
        \cline{2-2}
        & \textbf{to,like,would,you,leave},?,what,:,time,arrive\\

\end{tabular}
    \caption{Nearest 3 dialogue pairs and top 10 words corresponding to the vertices in the 3-D MVS-type polytope. Key semantic patterns in the dialogue pairs are italicized, while the corresponding top words are marked in bold.}    
    \label{tab:3d}
\end{table*}


%
Figure \ref{fig:3d-mvs} presents the resulting 3D polytope. Notably, a subset of utterances align closely with the polytope’s contour, while the majority tend to aggregate loosely around its center. Hence, generating a minimum volume polytope is pivotal, ensuring the vertices are as proximate to most utterances as feasible. Subsequently, the four vertices on the 3D polytope could be analyzed through two methods:

\begin{enumerate}
    \item \textbf{Examining the top $k$ words of vertices.} This is determined based on the previously mentioned vertex coordinates of the polytope, with $k=10$.
    \item \textbf{Identifying Nearest Dialogue Pairs to Vertices.} This is gauged by Euclidean distance in the reduced 3-D subspace. The proximity of similar dialogue pairs, showcasing analogous semantic structures in the original bag-of-words subspace, provides insights into semantic structures linked to the vertices. Three nearest dialogue pairs for each vertex are considered.
\end{enumerate}

The insights derived from these approaches are summarized in Table  \ref{tab:3d}. Words integral to dialogue turn updates, e.g. \textit{from, deparing, where} indicating the slot \textit{train-departure}, are prominently featured in the top words for vertices V1-V4, and the related semantic structures, including \textit{... departing from \textsc{city-name}}, are present in the nearest dialogue pairs. We take note that dialogue state updates usually adhere to fixed syntactic patterns, with values often positioned in predetermined locations, marked in smaller capital letters within delexicalized values in Figure \ref{tab:100d}.

Similarly extracted semantic structures can be found among the top words for V3, V4, respectively: \textit{leaves at \textsc{time}}, and \textit{i need a train to \textsc{city-name}}. 


\renewcommand\tabularxcolumn[1]{m{#1}}
\newcolumntype{C}{>{\centering\arraybackslash}X}
\begin{table}[ht]
\centering
\begin{tabularx}{\columnwidth}{|>{\hsize=.1\hsize}C|>{\hsize=.55\hsize}C|>{\hsize=.35\hsize}C|}
\hline
\textbf{Vertex}	& \textbf{Top 10 words}	& \textbf{Extracted semantic patterns}\\
\hline
V0 & that, leaves, train, arrive, work, at, depart, is, want, can & depart/leave at \textsc{time}\\
V7 & on, day, sunday, leave, thursd, friday, saturd, monday, wednes, tuesda & on Monday | Tuesday | ...\\
V11 & for, lookin, am, you, book, people, train, m, leave, it & book for \textsc{number} people \\
V17 & by, arrive, :, time, 30, 45, in, 15, or, to & arrive by \textsc{time}\\
V18 & ', ', yes, please, day, hi, sure, hello, you, ok, okay & (Greeting words)\\
... & ... & ... \\
\hline
\end{tabularx}
\caption{Examples of top words and extracted semantic patterns in 25-D MVS.}
\label{tab:100d}
\end{table}

For V1, its nearest dialogue pairs are sentences with many tokens out of vocabulary of BERT and CPM, marked as "[UNK]", which are captured by vertex V1. We also note words related to multiple patterns present in the topword of V1, including \textit{leaves at \textsc{time}}, \textit{book for \textsc{number} people}. Following our previous work in researching best practices of using CPM, where we found that increasing the dimensionality of reduced subspace leads to vertices capturing structures with more quantity and granularity, we increase the number of vertices to 26, i.e. finding MVS on a 25-D polytope, and examine the top words again. We note that these semantic patterns related to V1 are now separated into multiple vertices V0 and V11, as shown in table \ref{tab:100d}. Additionally, the emergence of new patterns, including "\textit{arrive by \textsc{time}}" and a collection of greeting words, comes along with the increment in vertices. As a side note, tokens in Table \ref{tab:100d} are from tokenization of original text by WordPiece tokenizer, in order to align with the tokenization practice in later experiments. It is hypothesized that this enhancement in representation granularity by a greater vertex count will amplify performance--a claim substantiated in subsequent experiment sections.

\subsection{Modification to TripPy encoder}
The enhancements made to the BERT encoder within TripPy align with the principles of Context-Guided BERT. This methodology postulates that integrating contextual information within the computation of latent representation in self-attention networks (SAN) strengthens dependencies across neural representations, leading to enhanced performance.

In the previous section, we note that features extracted by CPM maintain a close association with the semantic structures within dialogue turns. We thus hypothesize that merging such features into the attention layer potentially directs the attention mechanism towards tokens important to dialogue states.

Traditionally, within BERT's attention layer hosting  $H$ heads, an input sequence with length $n$ are transformed to query $\matr{Q} ^ h \in \mathbb{R}^{n \times d}$ and key $\matr{K} ^ h \in \mathbb{R}^{n \times d}$ at the $h$-th head. $d$ is the dimension of query and key vectors. Standard self-attention at head $h$ $\matr{A}^h$ is calculated by 

\begin{equation}
    \matr{A}^h = \text{softmax}(\frac{\matr{Q} ^ h {\matr{K} ^ h} ^ T }{\sqrt{d_h}})
\end{equation}

where $\sqrt{d_h}$ is a scaling factor.

We fuse CPM coefficients into self attention layer by amalgamating the coefficients with query and key vectors:
\begin{equation}
    \begin{bmatrix}
        \hat{\matr{Q}}^h \\
        \hat{\matr{K}}^h
    \end{bmatrix} = (1 - \begin{bmatrix}
        \lambda_Q^h \\
        \lambda_K^h
    \end{bmatrix}) \begin{bmatrix}
        {\matr{Q}}^h \\
        {\matr{K}}^h
    \end{bmatrix} + \begin{bmatrix}
        \lambda_Q^h \\
        \lambda_K^h
    \end{bmatrix} \mathbf{a} \begin{bmatrix}
        \mathbf{U}_Q^h \\
        \mathbf{U}_K^h
    \end{bmatrix})
\end{equation}

where $\mathbf{a} = [a_1, a_2, \dots, a_{R+1}]$ represents the CPM vertex coefficients for the input sequence. The scalar weight $\lambda_Q^h, \lambda_K^h$, modulating the influence of context on the output and are dynamically adapted per training instances, are also learnt and adjusted according to training samples:

\begin{equation}
    \begin{bmatrix}
        \lambda_Q^h \\
        \lambda_K^h
    \end{bmatrix} = \tanh {\begin{bmatrix}
        {\matr{Q}}^h \\
        {\matr{K}}^h
    \end{bmatrix} \begin{bmatrix}
        \mathbf{V}_Q^h \\
        \mathbf{V}_K^h
    \end{bmatrix} + \mathbf{a} \begin{bmatrix}
        \mathbf{U}_Q^h \\
        \mathbf{U}_K^h
    \end{bmatrix} \begin{bmatrix}
        \mathbf{V}_Q^a \\
        \mathbf{V}_K^a
    \end{bmatrix} }
\end{equation}


$\mathbf{U}_Q^h, \mathbf{U}_K^h, \mathbf{V}_Q^h, \mathbf{V}_K^h (h\in1, \cdots, H)$, $\mathbf{V}_Q^a, \mathbf{V}_K^a$ are linear weight matrices learnt at fine-tuning stage. 

The output of attention layer $\hat{\matr{A}}^h$ is then calculated using the new key and query matrices, followed by overlaying CPM attention on top:

\begin{equation}
    \hat{\matr{A}}^h = \text{softmax}(\frac{\hat{\matr{Q}}^h \hat{\matr{K}}^{h ^ T} }{\sqrt{d_h}}) + 0.5 (\lambda_Q^h + \lambda_K^h) \hat{M}
\end{equation}

where $\hat{M} = \text{softmax} (M'), M'$ indicates the similarity between tokens in the same utterance, as defined in Equation 2 in \ref{sec:CPM}. We note that the nature of $\tanh$ function ensures that $\lambda_Q^h, \lambda_K^h \in [-1,1]$, enabling the CPM context to augment attention both positively and negatively. This enhancement, albeit affecting the convexity properties of attention, has been validated in preceding studies to enrich the representation space of attention matrices in other NLP tasks \cite{wu_topic_2018}. We conjecture that these enhancements, demonstrated to be related to hints for dialogue processing, would have favorable implications in DST, which will be assessed in the experiment section below.

\section{Experiments}

\subsection{Datasets}

To evaluate the proposed model, we use MultiWoZ dataset, which is a dialogue dataset for DST task composed of dialogues in multiple domains: \textit{train, restaurant, hotel, taxi, }and\textit{attraction} with 30 domain-slot pairs in total. We use DST performance on this dataset to validate our previous claim that CPM-enhanced model can perform better on DST. 

Since MultiWoZ version 2.1 \cite{eric_multiwoz_2020}, which is the most up-to-date version published by the original MultiWoZ author, contains many annotation mistakes across training and test subset. Multiple attempts have been published to fix these mistakes partially, including MultiWoZ 2.2 \cite{zang_multiwoz_2020}, 2.3 \cite{han_multiwoz_2021}, 2.4 \cite{ye_multiwoz_2022}, and we report our result across all four versions to mitigate the effect of annotation error on our result, as published DST approaches are reported to score different across versions \cite{ye_multiwoz_2022}. We follow original train-val-test split published by authors of respective revisions of MultiWoZ, where validation and test splits each contain 1,000 dialogues and train splits have 8,438. Dialogues are not re-shuffled across trials on different MultiWoZ versions.

\subsection{Evaluation}

To evaluate DST performance, joint goal accuracy (JGA), the percentage of correctly predicted dialogue states \textit{at the last turn of dialogue} among all test dialogues, is used. Given that multiple works have reported instability of result over repeated trials, all results are averaged over 3 times with different seed. All results are assessed using two-sample t-test with $p<0.05$ recognized as statistically significant.

\subsection{Implementation}
The dataset is tokenized using WordPiece tokenizer, and the input sequence is trimmed to 512 tokens. After tokenization, we apply CPM algorithm using only current turn of user input and the immediately preceding system utterance, on training and development splits combined. For the fine-tuning stage, we mostly follow the original configuration of TripPy, which uses a pre-trained BERT model (BERT-base) for uncased text. All classification heads are randomly initialized and trained alongside the fine-tuning process of BERT on our selected training dataset. We fine-tune a different BERT for each version of MultiWoZ considering annotation discrepancy. The same CPM vertices set is used across different versions as CPM is label-agnostic and independent, and original dialogues remain identical in all four versions.
Training loss is calculated as a combination of cross entropy loss from slot action classifier $\mathcal{L}_\text{slot}^s$, span predictor $\mathcal{L}_\text{start}^s, \mathcal{L}_\text{end}^s$ and selector from system informed value memory and dialogue state memory $\mathcal{L}_\text{sysinform}^s, \mathcal{L}_\text{ds}^s$:

\begin{gather}
\begin{gathered}
\begin{aligned}
\mathcal{L}^s = 0.8 \mathcal{L}_\text{slot}^s +0.05(\mathcal{L}_\text{start}^s \\
+ \mathcal{L}_\text{end}^s)+ 0.05(\mathcal{L}_\text{sysinform}^s+ \mathcal{L}_\text{ds}^s)
\end{aligned}
\end{gathered}
\end{gather}

If in the entire dialogue there is no incidence of slot values originating from informed value and dialogue state, then the loss for that dialogue would be:

\begin{equation}
    \mathcal{L}^s = 0.8 \mathcal{L}_\text{slot}^s +0.1(\mathcal{L}_\text{start}^s + \mathcal{L}_\text{end}^s)
\end{equation}

We select weights on subtask loss in Equation 7 and 8 empirically. The loss of span predictor is set to 0 when the related slot is categorized into \textit{none} or \textit{dontcare}. Total loss is the summation of all per slot losses with equal weight, which is used as the optimization target for AdamW optimizer. The subspace and representation points in CPM remain fixed throughout the fine-tuning stage.

The learning rate of BERT $\alpha$, learning rate of slot classifier and span predictor $\beta$, are tuned during hyperparameter selection on development split, as well as epoch at which early stopping is applied. Other configurations are kept identical to those in the original TripPy experiment.

\section{Results}

Our main result is presented in Table \ref{cpm_result}, where we note a statistically significant advantage for our CPM-enhanced BERT on MultiWoZ version 2.3 and 2.4. We regard the result of previous versions, on which our proposed pipeline performed similarly against baseline, as partially affected by annotation error, an observation echoed by the authors of latter revisions of MultiWoZ when testing different DST models, in which all of the models suffered performance drag due to annotation error. \cite{ye_multiwoz_2022} While CPM algorithm is annotation-agnostic, the evaluation result is subject to the accuracy of annotation on test sets, hence we consider the results on latter versions of datasets more accurate than earlier versions. 

As a cross-check of result validity and evidence of computational efficiency, Table \ref{cpm_eff} further verifies that the superior result achieved by our CPM-enhanced pipeline uses less time during training to reach convergence, which proves that the performance discrepancy between the proposed pipeline and baseline cannot be owed to more training time. Additionally, considering that CPM algorithm runs solely on CPU and takes around 5 minutes to process the whole training and development set of MultiWoZ corpus on an Intel i7 CPU, the computation cost of CPM algorithm can be considered as minimal compared to savings in training time on GPU, highlighting the resource efficiency factor of our proposed pipeline.

We follow up the main result by a set of ablation studies conducted on MultiWoZ 2.4, aimed at dissecting the contributions of CPM coefficient and CPM attention towards the semantic structuring and resulting performance enhancement. In these trials, we isolated and evaluated the impacts of CPM coefficient and CPM attention, reinitializing and training the model from the ground up for each. The ablation result in Table \ref{cpm_ablate}, echoes our initial hypothesis. The implementation of only the CPM coefficient marginally surpasses the baseline, albeit without attaining statistical significance, underscoring the synergistic value of incorporating both feature sets for maximized efficacy.

\begin{table}
\centering
\begin{tabular}{|c|c|c|c|c|}
\hline
    MultiWoZ version & v2.1 & v2.2 & v2.3 & v2.4 \\
     \hline
    Baseline & 55.3 & 56.0 & 63.0 & 64.75 \\
    \hline
    CPM-assisted model & \textbf{54.65} & 55.47 & \textbf{63.47} & \textbf{66.16} \\ 
    \hline
\end{tabular}
    \caption{Joint Goal Accuracy (\%) of DST on 4 version of MultiWoZ datasets of the baseline model, and knowledge injected model. Results in bold are statically significant.}
    \label{cpm_result}
\end{table}
\begin{table}
\centering
\begin{tabular}{|c|c|c|c|c|}
\hline
    MultiWoZ version & v2.1 & v2.2 & v2.3 & v2.4 \\
     \hline
    Baseline & 26 & 30.33 & 24.67 & 20.33 \\
    \hline
    CPM-assisted model & 23.33 & 29 & 20.67 & 16.33 \\ 
    \hline
\end{tabular}
    \caption{Average Training iteration used of DST on 4 version of MultiWoZ datasets of the baseline model, and knowledge injected model. All experiments were run on the same hardware setup over 3 different random seeds.}
    \label{cpm_eff}
\end{table}
\begin{table}
\centering
\begin{tabular}{|c|c|}
\hline
     & JGA \\
     \hline
    Baseline & 64.75 \\
    \hline
    CPM attention only & 63.70 \\
    \hline
    CPM coefficient only & 64.76 \\
    \hline
    CPM-assisted model & \textbf{66.16} \\
    \hline
\end{tabular}
    \caption{Ablation study result of CPM assisted model, measured by JGA (\%) 
 on MultiWoZ 2.4 dataset. Result in bold is statistically significant. }
    \label{cpm_ablate}
\end{table}
\section{Discussion}

\begin{table}
\centering
\begin{tabular}{|c|c|c|c|c|}
\hline
    CPM dimention & 10 & 25 & 50 & 100 \\
     \hline
    JGA & 64.45 & 64.60 & 65.40 & 66.16 \\ 
    \hline
\end{tabular}
    \caption{Joint Goal Accuracy (\%) of DST on CPM-assisted TripPy with different dimensionality. Only MultiWoZ 2.4 dataset is used in this experiment.}
    \label{cpm_dim}
\end{table}

\subsection{Effect of Convex Polytope Dimensionality}
As elaborated in Section \ref{sec:CPM}, the dimensionality of the CPM polytope is directly correlated to the intricacy of resolved semantic patterns within the corpus. Considering the extensive and varied linguistic expressions inherent in MultiWoZ 2.4, it is postulated that higher-dimensional configurations of CPM would be better in optimizing DST performance by providing a richer, more granular semantic structure for the neural model to learn from. This theoretical positioning is substantiated by the data presented in Table \ref{cpm_dim}, a direct positive correlation between the dimensionality of the CPM model's convex polytope and the DST performance.

\begin{figure*}[!thb]
    \centering
    \includegraphics[width=1\textwidth,height=\textheight,keepaspectratio]{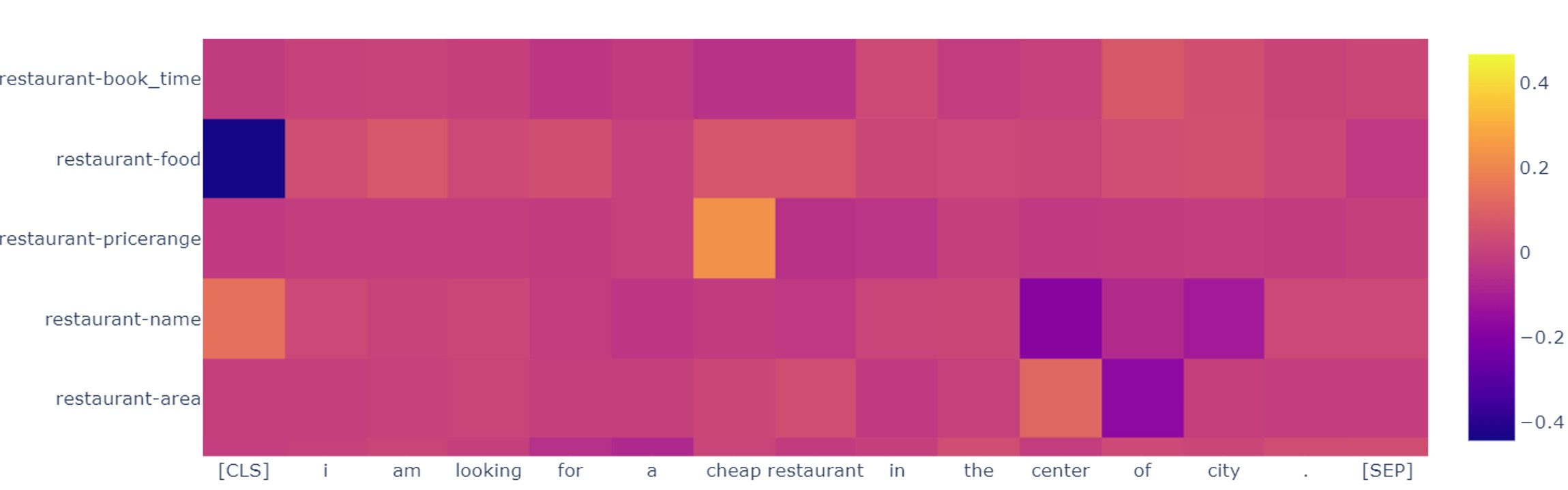}
    \caption{Normalized change of Integrated Gradient (IG) between an input sequence and slot action prediction, calculated as IG on CPM-assisted TripPy subtracted by IG on vanilla TripPy. Only a few slots are displayed due to size constraints. Higher value of IG indicates more positive attribution.}
    \label{fig:attdelta}
\end{figure*}
\begin{figure*}[!htb]
    \centering
    \includegraphics[width=0.98\textwidth,height=\textheight,keepaspectratio]{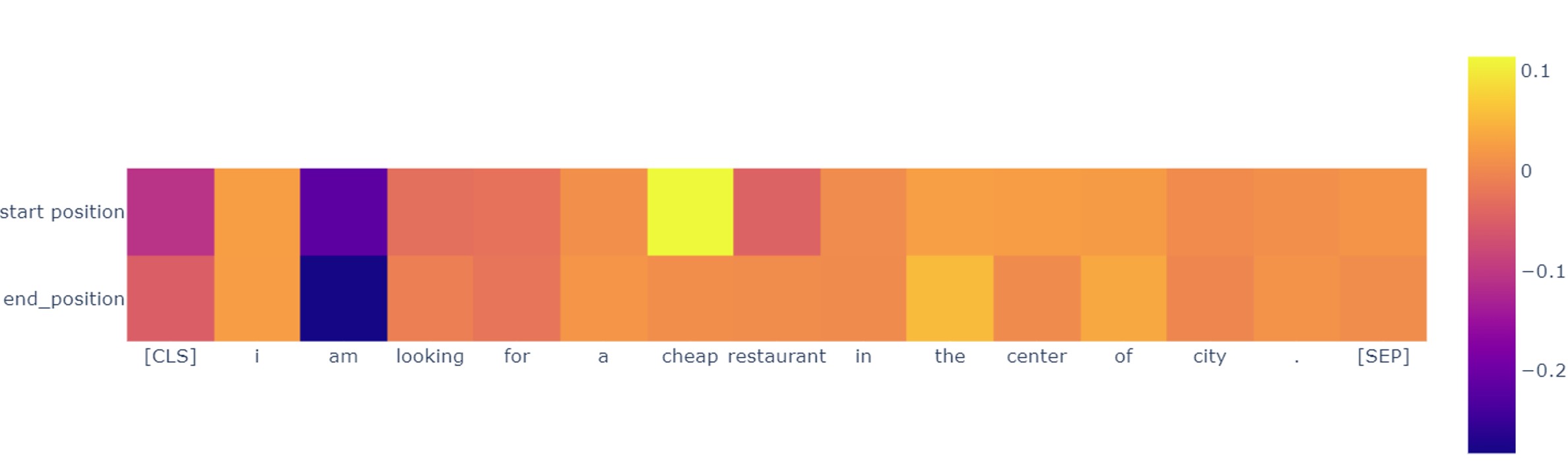}
    \caption{Normalized change of Integrated Gradient (IG) between an input sequence and span prediction for slot \textit{restaurant-pricerange} on CPM-assisted TripPy, compared to vanilla TripPy. Higher value of IG indicates more positive attribution.}
    \label{fig:span}
\end{figure*}
\begin{figure*}[!htb]
    \centering
    \includegraphics[width=0.98\textwidth,height=\textheight,keepaspectratio]{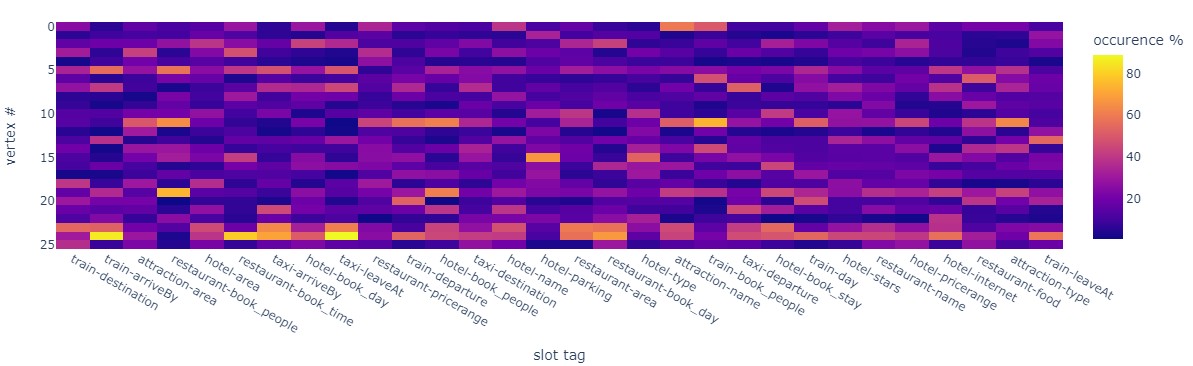}
    \caption{Occurrence of vertex in the list of most important vertices with respect to individual slots, normalized by dividing against total occurrence of individual slots. Higher occurrence percentage indicates more significance.}
    \label{fig:freq}
\end{figure*}

\subsection{Influence of CPM features on BERT}

Beyond the empirical performance analysis, a qualitative exploration has been undertaken to ascertain the precise manner in which CPM influences the inference process within the modified TripPy model. We theorize that the incorporation of CPM features would enhance the correlation between correct dialogue state predictions and the relevant key tokens within the input sequence.

Integrated Gradients (IG), a method surpassing vanilla gradient calculations in reproducibility and sensitivity, has been employed for this analysis \cite{DBLP:conf/corr/SundararajanTY17}. With the model's weights held constant, IG visualizes the attributions between input tokens and \textbf{correct} action predictions on a per-slot basis. An inference mode analysis illustrated that tokens with higher IG values held a more significant influence on correct predictions, depicted as lighter color for a higher value when drawing Figure \ref{fig:attdelta} and \ref{fig:span}. Subsequent normalization and comparison of IG values between the CPM-assisted and baseline models reveal the net changes in attributions, providing insights into CPM's role in refocusing model attention on contextually crucial tokens, aligning more with human interpretational logic.

We compute normalized IG of slot action prediction result against input sequence \textsc{[CLS] i am looking for a cheap restaurant in the center of city . [SEP]} in both our CPM-enhanced pipeline and baseline model, and plot the delta in Figure \ref{fig:attdelta}. The result emphasizes the relatively heightened attribution to the token "CHEAP" within the sequence, when predicting the \textit{restaurant-pricerange} slot using the CPM-assisted model, compared with baseline. This elevated focus on contextual elements is consistent with human linguistic understanding, emphasizing the pivotal role of specific tokens like "cheap" in determining dialogue states for relevant slots. As a cross-check, Figure \ref{fig:span} about normalized change of IG for span prediction subtask on our pipeline also shows that aside from slot action prediction, the token "CHEAP" is also correctly attributed with more significance for span start in span prediction subtask compared to baseline, consistent with our findings in Figure \ref{fig:attdelta}.

Besides the analysis of elevated attribution on key input tokens thanks to CPM, we also directly computed IG between CPM coefficient and the correct prediction of slot actions. The preliminary analysis on \textit{train} domain in Table \ref{tab:3d} and \ref{tab:100d} has already demonstrated that vertex coefficients in CPM model are closely related to semantic structures in the collection of utterances, and the overlap of semantic structures between different vertices goes smaller as the dimensionality increases. Therefore, vertices coefficients should be assigned with varying attributions across different slots. To obtain a more conclusive answer, for each input sequence in the test set whose prediction of slot actions is not entirely \textit{none}, i.e. there is at least one pending update to the dialogue state after this dialogue turn, the IG between CPM coefficient and correct \textbf{not} \textit{none} prediction of slot action is computed, and for a slot with \textbf{not} \textit{none} action, 5 vertices with the largest attribution are designated as \textbf{important vertices} with respect to the slot. The frequency of a vertex as the important vertex to a particular slot is summarized across the whole test set, and illustrated in Figure \ref{fig:freq}, where it is clear that different vertices contribute to the correct predictions of different slots. Exceptions are vertex \#23, 24 displayed in the bottom, who are related to many slots. Their top-words include \textit{your, help, need, all} and \textit{at, on, time, by}, which are common words in the corpus, and may not display affinity to particular slot domains or attributes. By contrast, the occurrence pattern of other vertices as important vertex is different across different slots. This further validate our argument that CPM features are indeed semantic-related, and could be interpreted using the visualization techniques introduced in section \ref{sec:CPM}, and the rationale of enhanced performance in the CPM-assisted DST model could be explained by the interpretation of extracted semantic structure by CPM.


\section{Conclusion}

This paper introduces a framework designed to integrate linguistic knowledge into BERT models, aiming to elevate performance in Dialogue State Tracking (DST) within task-oriented dialogues, without additional training data, annotation, and excessive computational resources. The integration of linguistic knowledge, extracted without supervision through the geometric Convex Polytope Model (CPM), offers insights that are interpretable both geometrically and through feature attribution, allowing an in-depth understanding of the correlation between geometric properties and predictive outcomes. This inclusion of external knowledge into the BERT model as input leads to enhancements in performance compared to the baseline, with minimal additional computation resource expense. highlighting the efficacy of the proposed framework. Under real life conditions, our pipeline can accommodate new data samples as CPM algorithm is lightweight and runs quick, and being label-agnostic make the feature extraction process robust against noises in annotation. We leave detailed exploration on training NLP models under suboptimal labelling quality and/or quantity as future work.

While this paper uses a BERT-based model to showcase the virtue of transparency brought into DST neural model from CPM, future endeavors will explore the incorporation of this framework into more sophisticated neural models and will probe into the relational dynamics of CPM-extracted features with super-sentence level relationships, e.g. semantic themes within paragraphs, extending the applicability of this framework to a range of NLP tasks. The combination of linguistic understanding and complex modelling in NLP tasks presents a direction towards visibility and confidence in the decision process of increasingly complex NLP models, which lays the foundation of mass trust of adoption.

\renewcommand{\bibfont}{\referencefont\vskip 0.3\baselineskip plus 0.1\baselineskip minus 0.1\baselineskip}
\bibliography{naacl,paper}
\bibliographystyle{IEEEtran}


\begin{IEEEbiographynophoto}{Xiaohan Feng} received his B.Eng (Hons.) degree in Information Engineering from the Chinese University of Hong Kong, Hong Kong, in 2016. He is currently a PhD candidate at Human-Computer Communications Lab, Chinese University of Hong Kong, Hong Kong. His research interests include natural language processing and interpretable AI. 
\end{IEEEbiographynophoto}
\begin{IEEEbiographynophoto}{Xixin Wu} (Member, IEEE) received the B.S. degree from Beihang University,
Beijing, China, the M.S. degree from Tsinghua University, Beijing, and the
Ph.D. degree from The Chinese University of Hong Kong, Hong Kong. He is
currently an Assistant Professor with the Department of Systems Engineering
and Engineering Management, The Chinese University of Hong Kong. Before
this, he was a Research Associate with the Machine Intelligence Laboratory, Engineering Department, Cambridge University, Cambridge, U.K., and a Research
Assistant Professor with the CUHK Stanley Ho Big Data Decision Analytics
Research Centre. His research interests include speech synthesis and recognition,
speaker verification, and neural network uncertainty. 
\end{IEEEbiographynophoto}
\begin{IEEEbiographynophoto}{Helen Meng} (Fellow, IEEE) received the B.S., M.S., and Ph.D. degrees in electrical engineering from the Massachusetts Institute of Technology, Cambridge,
MA, USA. In 1998, she joined the Chinese University of Hong Kong, Hong
Kong, where she is currently the Chair Professor with the Department of Systems
Engineering \& Engineering Management. She was the former Department
Chairman and the Associate Dean of Research with the faculty of Engineering.
Her research interests include human–computer interaction via multimodal and
multilingual spoken language systems, spoken dialog systems, computer-aided
pronunciation training, speech processing in assistive technologies, health related applications, and Big Data decision analytics. She was an Editor-in-Chief
of the \textit{IEEE TRANSACTIONS ON AUDIO, SPEECH AND LANGUAGE PROCESSING}
between 2009 and 2011. She was the recipient of the IEEE Signal Processing
Society Leo L. Beranek Meritorious Service Award in 2019. She was also on the
Elected Board Member of the International Speech Communication Association
(ISCA) and an International Advisory Board Member. She is a fellow of ISCA, HKCS,
and HKIE.
\end{IEEEbiographynophoto}
\EOD

\end{document}